\documentclass[letterpaper, 10 pt, conference]{ieeeconf}  

\IEEEoverridecommandlockouts                              %

\usepackage{algorithmic}
\usepackage{algorithm}
\usepackage{array}
\usepackage{textcomp}
\usepackage{stfloats}
\usepackage{verbatim}
\usepackage{cite}
\usepackage{fancyhdr}

\usepackage[para]{threeparttable}

\usepackage{booktabs}

\usepackage{arydshln}
\usepackage{gensymb}

\usepackage{subfigure}

\usepackage[hyphens,spaces,obeyspaces]{url}

\usepackage{amsmath,amsfonts,amssymb}

\usepackage[pdftex]{graphicx}

\usepackage{tikz}

\usepackage{xcolor}
\usepackage{xparse,xcoffins}

\ExplSyntaxOn
\NewCoffin\imagecoffin
\NewCoffin\labelcoffin

\keys_define:nn { miguel/label }
 {
  label   .tl_set:N = \l_miguel_label_tl,
  labelbox .bool_set:N = \l_miguel_label_box_bool,
  labelbox .default:n = true,
  fontsize .tl_set:N = \l_miguel_label_size_tl,
  fontsize .initial:n = \footnotesize,
  pos .choice:,
  pos/nw .code:n = \tl_set:Nn \l_miguel_label_pos_tl { left,up },
  pos/ne .code:n = \tl_set:Nn \l_miguel_label_pos_tl { right,up },
  pos/sw .code:n = \tl_set:Nn \l_miguel_label_pos_tl { left,down },
  pos/se .code:n = \tl_set:Nn \l_miguel_label_pos_tl { right,down },
  pos/n .code:n = \tl_set:Nn \l_miguel_label_pos_tl { hc,up },
  pos/w .code:n = \tl_set:Nn \l_miguel_label_pos_tl { left,vc },
  pos/s .code:n = \tl_set:Nn \l_miguel_label_pos_tl { hc,down },
  pos/e .code:n = \tl_set:Nn \l_miguel_label_pos_tl { right,vc },
  pos .initial:n = nw,
  unknown .code:n   = \clist_put_right:Nx \l_miguel_label_clist
                       { \l_keys_key_tl = \exp_not:n { #1 } }
 }
\clist_new:N \l_miguel_label_clist
\box_new:N \l_miguel_label_box
\box_new:N \l_miguel_label_image_box

\NewDocumentCommand{\xincludegraphics}{O{}m}
 {
  \group_begin:
  \tl_clear:N \l_miguel_label_tl
  \clist_clear:N \l_miguel_label_clist
  \keys_set:nn { miguel/label } { #1 }
  \tl_if_empty:NTF \l_miguel_label_tl
   {
    \miguel_includegraphics:Vn \l_miguel_label_clist { #2 }
   }
   {
    \SetHorizontalCoffin\imagecoffin
     {
      \miguel_includegraphics:Vn \l_miguel_label_clist { #2 }
     }
    \SetHorizontalCoffin\labelcoffin
     {
      \raisebox{\depth}
       {
        \bool_if:NTF \l_miguel_label_box_bool
         { \fcolorbox{black}{white}{\l_miguel_label_size_tl\l_miguel_label_tl} }
         { \l_miguel_label_size_tl\l_miguel_label_tl }
       }
     }
    \SetVerticalPole\imagecoffin{left}{3pt+\CoffinWidth\labelcoffin/2}
    \SetVerticalPole\imagecoffin{right}{\Width-3pt-\CoffinWidth\labelcoffin/2}
    \SetHorizontalPole\imagecoffin{up}{\Height-3pt-\CoffinHeight\labelcoffin/2}
    \SetHorizontalPole\imagecoffin{down}{3pt+\CoffinHeight\labelcoffin/2}
    \use:x{\JoinCoffins\imagecoffin[\l_miguel_label_pos_tl]\labelcoffin[vc,hc]} 
    \TypesetCoffin\imagecoffin
   }
   \group_end:
 }
\NewDocumentCommand{\setlabel}{m}
 {
  \keys_set:nn { miguel/label } { #1 }
 }

\cs_new_protected:Nn \miguel_includegraphics:nn
 {
  \includegraphics[#1]{#2}
 }
\cs_generate_variant:Nn \miguel_includegraphics:nn { V }

\ExplSyntaxOff

\fancypagestyle{IEEEtitlepagestyle}{
  \fancyhf{} 
  \fancyhead[L]{\small\textcolor{gray} {Appears in IEEE International Conference on Robotics and Automation (ICRA), London, 2023}}
}

\begin{document}

\title{BogieCopter: A Multi-Modal Aerial-Ground Vehicle \\for Long-Endurance Inspection Applications}
\author{ Teodoro Dias$^{1}$, Meysam Basiri$^{*1}$
\thanks{$^{1}$Teodoro Dias and Meysam Basiri are with Institute for Systems and Robotics of the Instituto Superior Técnico, Universidade de Lisboa, Portugal
        . This work
was supported by the strategic funding LARSy through Project FCT under Grant
UIDB/50009/2020. {\tt\small meysam.basiri@tecnico.ulisboa.pt}}%
}



\maketitle
\thispagestyle{IEEEtitlepagestyle}


\begin{abstract}

The use of Micro Aerial Vehicles (MAVs) for inspection and surveillance missions has proved to be extremely useful, however, their usability is negatively impacted by the large power requirements and the limited operating time. This work describes the design and development of a novel hybrid aerial-ground vehicle, enabling multi-modal mobility and long operating time, suitable for long-endurance inspection and monitoring applications. The design consists of a MAV with two tiltable axles and four independent passive wheels, allowing it to fly, approach, land and move on flat and inclined surfaces, while using the same set of actuators for all modes of locomotion. In comparison to existing multi-modal designs with passive wheels, the proposed design enables a higher ground locomotion efficiency, provides a higher payload capacity, and presents one of the lowest mass increases due to the ground actuation mechanism. The vehicle’s performance is evaluated through a series of real experiments, demonstrating its flying, ground locomotion and wall-climbing capabilities, and the energy consumption for all modes of locomotion is evaluated.

\end{abstract}





\section{Introduction}


During the last decade, Micro Aerial Vehicles (MAVs) have received a great deal of attention, both academically and commercially, due to their ability to quickly reach areas of interest, overcome obstacles, and provide an elevated view of the environment. The applications in which MAVs are used have been multiplying and include: 
precision agriculture \cite{agriculture}, inspection \cite{inspection}, package delivery \cite{delivery}, search and rescue \cite{basirijfr, disaster} and surveillance \cite{pegasus}. 
Despite the clear benefits of MAVs for such applications, their use is heavily impacted by the short operating time 
 due to the high power requirements needed for flying and the limited payload capacity. 

One way to mitigate these limitations is by considering multi-modal MAV designs that allow a better utilization of energy for different tasks and environments \cite{energeticsDesign}. 
Multi-modal aerial-ground vehicles combine the ability of aerial locomotion with the ability of ground locomotion in a single hybrid vehicle. It allows incorporating some of the benefits of a ground vehicle (GV) into the MAV, such as the ability to perform low powered navigation and the ability to move in confined spaces. 
Multi-modal vehicles can allow operation in complex environments that need different locomotion capabilities, and to significantly increase the operation range and time of the vehicle \cite{HyTAQ2, rollocopter_v2, monoWheeled}. 
Such vehicles have been gaining attention in applications where flight is desirable to rapidly identify a problem or a target and ground locomotion is desirable to perform the majority of the task, maximizing the vehicle's operating time \cite{MiguelPaper, drivocopter, cagedTwoWheels}.



\begin{figure}[t]
\centering
\includegraphics[width=0.59\linewidth]{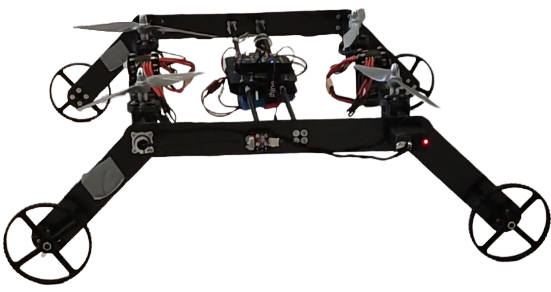}
\vspace{-0.5\baselineskip}
\caption{BogieCopter: A multi-modal micro aerial-ground vehicle}
\label{fig:bogiecopter}
\end{figure}

This work presents an innovative, four-wheeled single-axis dual-tiltable bimodal MAV, named BogieCopter, presented in figure \ref{fig:bogiecopter}, capable of efficient aerial, ground and surface locomotion and suitable for long-endurance inspection and monitoring applications. To design this vehicle,  
an extensive review of the state of the art in multi-modal MAVs was initially conducted to identify the strength and limitations of existing designs. To the best of our knowledge, this is the most complete review of research available for multi-modal aerial-ground vehicles. This work aims at addressing some of the limitations in prior designs, by exploiting the strengths of both active and passive solutions offered for the ground actuation mechanism, while taking into consideration a generous payload required for most industrial applications. 
BogieCopter was designed from the ground up to have passive ground actuation mechanism (GAM), i.e. using passive wheels and relying on the same set of actuators for both aerial and ground locomotion. This results in a minimal impact of the GAM on the maximum take-off mass (MTOM). By using tiltable rotors, BogieCopter decouples the ground and the flight controller, achieves low dust generation, and is capable of efficient locomotion. Furthermore, the design takes into consideration the protection of the propellers, allowing safe navigation of the vehicle in proximity of obstacles and robustness to collisions. A validation and an analysis of power consumption is demonstrated for all modes of locomotion, demonstrating the advantage of the proposed design, and multi-modal aerial-ground vehicles in general, for performing long-endurance operations.  



The outline of the paper is as follows: Section \ref{section:stateoftheart} provides a detailed review of the state-of-the-art on multi-modal MAVs and describes the limitations of the existing passively actuated multi-modal MAVs.
Section \ref{section:bogieCopterDesign} describes the proposed multi-modal MAV design and the design procedure. 
Section \ref{section:results} presents results from real-world experiments that validate the design for aerial, ground, inclined surface and wall-climbing locomotion and provides an analysis of the power consumption of the vehicle for all modes of locomotion. Section \ref{section:conclusion} concludes the document and presents the future work.

\section{State-of-the-art}
\label{section:stateoftheart}


Prior work on multi-modal aerial-ground vehicles largely falls into two main categories of  \textbf{Passive actuated} or  \textbf{Active actuated} depending on the actuation mechanism used for the ground mode. In passive actuated designs, the same actuators used for flight enable the MAV to move on the ground, while in active actuated designs additional actuators are used, usually electric motors, for the ground locomotion. Table \ref{tab:summaryCharacteristics} presents all the prior work on this regard, both commercial and academic, as well as some characteristics that are of interest in multi-modal vehicles. It should be noted that, although not aiming for a hybrid aerial-ground vehicle, both \cite{wallClimber} and \cite{bi2copter} are included since they focus on the capability of the MAV to move in contact with walls and can be considered as a special case of multi-modality. By analysing table \ref{tab:summaryCharacteristics}, it is possible to conclude:

\begin{itemize}
    \item Multi-modal vehicles have the potential to be used in a variety of applications;
    \item Most of the designs are tailored toward a specific use case and cannot be used for other applications since low Thrust-to-Weight ratios (T/W ratio) are used or details such as the payload are not specified.
    \item With the T/W ratios used, a low safety factor (SF) is obtained, which prevents the vehicles to be considered safe to fly in windy conditions;
    \item Most vehicles have a low or no payload capacity at all, being unable to carry commonly used industrial sensors for inspection applications such as high precision thermal sensors or LIDARs;
    \item The advantages of a hybrid design for enhancing the vehicle's operating time can be observed. The addition of a GAM allows an increase of between 1.33 to 11.25 times in the operating time when ground locomotion is used instead of aerial locomotion. 
    \item The minimum mass achieved for the GAM in terms of the MTOM is 1.03\% by Gemini \cite{monoWheeled, gemini}. However, this design requires continuous actuation to stay static on the ground, having a high power consumption of 106 W. Considering usual designs that don't require energy to be stationary on a flat surfaces, the lowest mass attained for the GAM, with respect to the MTOM, is 11.8\%, by Rollocopter \cite{rollocopter_v2};
    \item  As expected, the GAM's impact on the vehicle's MTOM in active actuated designs is higher than the GAM's impact on the vehicle's MTOM for passive actuated designs.
    \item The active actuated designs exhibit a higher increase in the operation time when using ground locomotion compared to the passive designs. 
\end{itemize}

As evident from the literature, although passive actuated designs are more efficient during flight, due to their lower GAM mass, they are less efficient in ground mode when compared to active designs. 
This is because, passive actuated aerial-ground vehicles, with the exception of 
\cite{flyingSTAR}, have always followed a common design principle that results in a dependant and coupled dynamics and controllers for the ground and aerial locomotion modes, preventing the ability to optimize the ground locomotion. In addition such designs lead to poor stability and a high level of dust generation \cite{drivocopter}.
In the following subsection, we will analyse in more detail the approach used for conventional passive designs.

\begin{table*}[t]
\begin{minipage}[h]{1.0\linewidth}
\centering
\begin{threeparttable}
\vspace{-0.5\baselineskip}
\caption{Characteristics of interest of the different reviewed multi-modal rotary-wing MAVs}
\label{tab:summaryCharacteristics}
\renewcommand{\arraystretch}{1.1}
  \begin{tabular}{>{\centering}m{0.15\linewidth}                            >{\centering}m{0.1\linewidth}
                  >{\centering}m{0.1\linewidth}
                  >{\centering}m{0.1\linewidth}
                  >{\centering}m{0.07\linewidth}
                  >{\centering}m{0.085\linewidth}
                  >{\centering}m{0.095\linewidth}
             >{\centering\arraybackslash}m{0.09\linewidth}}
     \toprule
     \centering \textbf{MAV} & \centering \textbf{Main Use Case} & \centering \textbf{Payload} & \centering \textbf{Maximum Take-off Mass (MTOM)} & \centering \textbf{Thrust-to-Weight Ratio} & \centering \textbf{Type of GAM} & \centering \textbf{GAM's Mass compared to MTOM (\%) } & {\centering \textbf{Ground Operating Time\tnote{1}}} \\ 
     \midrule
     \centering Pegasus \cite{pegasus} & \centering Surveillance & \centering NS\tnote{2} & \centering NS & \centering NS & \centering Active & \centering NS 
     & {\centering NS} \\ 
     \hdashline
     \centering B-Unstoppable \cite{b-unstoppable} & \centering Hobby & \centering NS  & \centering NS  & \centering NS  & \centering Active & \centering NS  & {\centering $>$ 1.33x} \\
     \hdashline
     \centering \cite{morphTwo} & \centering NS  & \centering NS  & \centering 946 g\tnote{3}
     & \centering NS  & \centering Active & \centering NS  & {\centering NS} \\
     \hdashline
     \centering \cite{secondMorphTwo} & \centering Disaster Sites\tnote{4}  & \centering 110 g  & \centering 450 g
     & \centering $\sim$ 2.3\tnote{3} & \centering Active & \centering $\sim$ 19.8\tnote{5}  & {\centering $\sim$ 23.6} \\
     \hdashline
     \centering WAMORN \cite{WAMORN} & \centering Disaster Sites & \centering NS  & \centering 350 g
     & \centering $\sim$ 1.06 & \centering Active & \centering NS  & {\centering NS} \\
     \hdashline
     \centering \cite{quatro} & \centering Crop Evaluation & \centering NS & \centering NS  & \centering NS  & \centering Active & \centering NS 
     & {\centering $<$ 1.75x} \\
     \hdashline
     \centering \cite{wallClimber} & \centering Inspection & \centering NS  & \centering NS  & \centering NS  & \centering Active & \centering NS  & {\centering NS } \\
     \hdashline
     \centering MTMUR \cite{MTMUR} & \centering NS  & \centering NS & \centering 1.5 kg\tnote{3}
     & \centering $\sim$ 3.4  & \centering Active & \centering NS & {\centering NS } \\
     \hdashline
     \centering Drivocopter \cite{drivocopter} & \centering DARPA SubT Challenge & \centering 850 g & \centering 5.1 kg & \centering NS  & \centering Active & \centering $\sim$ 17.6 & {\centering $\sim$ 11.25x} \\
     \hdashline
     \centering JJRC H3 \cite{jjrcH3} & \centering Hobby & \centering \centering NS  & \centering NS  & \centering NS  & \centering Active & \centering NS  & {\centering NS } \\
     \hdashline
     \centering Syma X9 \cite{symaX9} & \centering Hobby & \centering \centering NS  & \centering NS  & \centering NS  & \centering Active & \centering NS  & {\centering NS } \\
     \hdashline
     \centering HyTAQ \cite{HyTAQ1, HyTAQ2} & \centering NS & \centering NS & \centering 570 g\tnote{3} & \centering $\sim$ 2.36 & \centering Passive & \centering NS & {\centering $\lesssim$ 6x} \\
     \hdashline
     \centering \cite{cagedTwoWheels} & \centering Bridge Inspection & \centering $\geq$ 74 g & \centering 1.36 kg\tnote{3} & \centering NS  & \centering Passive & \centering NS  & {\centering NS } \\
     \hdashline
     \centering MUWA \cite{MUWA} & \centering Disaster Sites & \centering \centering NS  & \centering 2.1 kg & \centering $\sim$ 2.16 & \centering Passive & \centering NS  & {\centering NS } \\
     \hdashline
     \centering PRSS UAV \cite{PRSS_UAV} & \centering Disaster Sites & \centering $\geq$ 76 g & \centering 2kg & \centering $<$ 1.5  & \centering Passive & 
     \centering $\sim$ 33.2 & {\centering NS } \\
     \hdashline
     \centering Rollocopter \cite{rollocopter_v1} & \centering Space Exploration & \centering NS  & \centering NS  & \centering NS  & \centering Passive & \centering NS  & {\centering NS } \\ 
     \hdashline
     \centering Shapeshifter 
     \cite{shapeshifter_2} & \centering Space Exploration  & \centering NS  & \centering $\sim$ 800 g & \centering $\sim$ 4.08 & \centering Passive & \centering NS  & {\centering NS\tnote{6}} \\
     \hdashline
     \centering Gemini \cite{monoWheeled, gemini} & \centering Confined Spaces & \centering 500 g & \centering 1.95 kg\tnote{3} & \centering $\sim$ 1.28 & \centering Passive & \centering $\sim$ 1.03 & {\centering $\lesssim$  2.58x} \\
     \hdashline
     \centering NINJA UAV \cite{ninjaUAV} & \centering NS  & \centering NS  & \centering 2.122 kg\tnote{3} & \centering NS  & \centering Passive & \centering NS  & {\centering NS } \\ 
     \hdashline
     \centering Rollocopter \cite{rollocopter_v2} & \centering NS  & \centering 850 g\tnote{7} & \centering 4.231 kg & \centering $\sim$ 1.47 & \centering Passive & $\sim$ 11.8 & {\centering $\sim$ 5} \\
     \hdashline
     \centering Parrot Rolling Spider \cite{rollingSpider} & \centering Hobby & \centering NS  & \centering NS  & \centering NS  & \centering Passive & \centering NS  & {\centering NS } \\ 
     \hdashline
     \centering Flying STAR \cite{flyingSTAR} & \centering Confined Spaces & \centering NS  & \centering 900 g\tnote{3} & \centering $\sim$ 1.36 & \centering Passive & \centering NS  & {\centering NS } \\ 
     \hdashline
     \centering Inkonova Tilt Scout \cite{inkonova} & \centering Subterranean Inspection & \centering 300 g  & \centering NS  & \centering NS  & \centering Passive & \centering NS  & {\centering NS } \\
     \hdashline
     \centering Quadroller \cite{quadroller} & \centering NS  & \centering NS  & \centering 1.3 kg\tnote{3} & \centering NS  & \centering Passive & \centering $\sim$ 22.3 & {\centering $\sim$ 2} \\
     \hdashline
     \centering Bi$^2$Copter \cite{bi2copter} & \centering Bridge Inspection & \centering $\sim$ 2.34 kg & \centering $\sim$ 4.84 kg & \centering $\sim$ 1.26 & \centering Passive & \centering NS  & {\centering NS } \\ 
     \bottomrule
     \end{tabular}
\begin{tablenotes}
\item [1] Increase over flight operating time
\item [2] Not Specified
\item [3] Unknown if it is higher 
\item [4] Main use case
\item [5] Weight of the shared components not considered as part of the GAM's mass
\item [6] Similar payload to \cite{drivocopter}
\item [7] Theoretical operating ranges given
\end{tablenotes}
\end{threeparttable}
\end{minipage}
\vspace{-1.5\baselineskip}
\end{table*}

\subsection{Conventional Passive Actuated Designs}
\label{subsection:passiveProblemsExplanation}

\begin{figure}[h]
\centering
\includegraphics[width=0.50\linewidth]{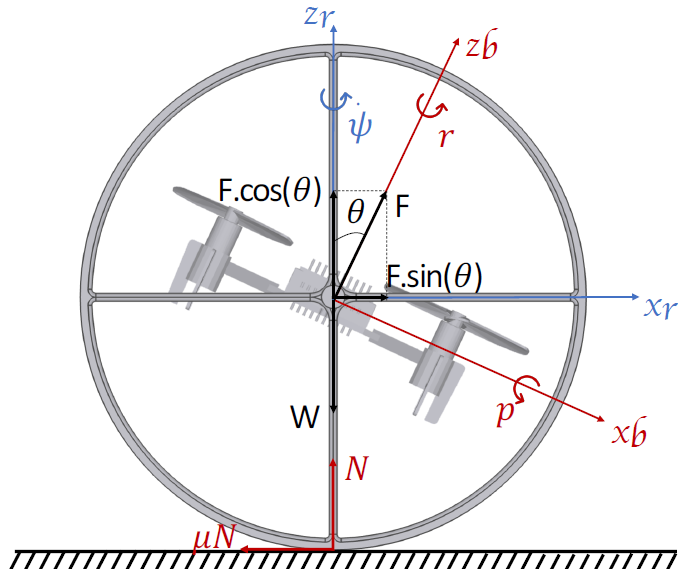}
\vspace{-0.5\baselineskip}
\caption{Conventional passive actuated design, with the longitudinal motion enabled by the pitching of the MAV \cite{MiguelPaper}}
\label{fig:conventionalPassive}
\end{figure}

Common passively actuated ground-aerial MAVs consist of a conventional quadrotor that is freely attached to the center of a cylindrical cage \cite{HyTAQ1, HyTAQ2}, a spherical cage \cite{rollocopter_v1, shapeshifter_2, MUWA, PRSS_UAV}, or a independent wheels' mechanism \cite{ninjaUAV, cagedTwoWheels, MiguelPaper, gemini, monoWheeled, rollingSpider, rollocopter_v2}, as shown in Figure \ref{fig:conventionalPassive}. Such designs achieve 
the longitudinal and lateral locomotion on the ground by controlling the attitude and thrust of the quadrotor.
While pitching, the thrust ($F$) produced by such MAVs can be decomposed into two forces, one perpendicular ($F_\perp$) and the other parallel ($F_\parallel$) to the ground: 
\begin{minipage}{0.96\linewidth}
\hspace*{\fill}
\begin{minipage}{.46\linewidth}
\centering
\begin{equation}
    F_\parallel = F \sin(\theta)
    \label{eq:parallelForce}
\end{equation}
\end{minipage}%
\hfill
\begin{minipage}{.46\linewidth}
\centering
\begin{equation}
    F_\perp = F \cos(\theta)
    \label{eq:perpendicularForce}
\end{equation}
\end{minipage}
\hspace*{\fill}
\end{minipage}

\vspace{6pt}
\noindent where 
$\theta$ is the pitch angle. 
It can be observed that only $F_\parallel$ is responsible for the longitudinal motion of the MAV. 
For maximum ground locomotion efficiency $F_\parallel$ should be maximized, implying that the vehicle must move at a pitch angle of 90$\degree$ to the ground. 
This can be generalized to inclined surfaces with slope angle $\psi$, where 
the pitch angle should be equal to:
\begin{equation}
    \theta = 90\degree - \psi
    \label{eq:inclinedSurface}
\end{equation}

The requirement to control the pitch for motions on the ground constrains the mechanical design and results in a similar and coupled controller for both ground and flight modes \cite{drivocopter}, preventing the optimization of each mode independently.
Furthermore, for an efficient locomotion on inclined surfaces, \textit{a priori} knowledge of the slope angle and adaptation of the pitch angle is required \cite{MiguelPaper}. The mechanical design constraint also limits the applicability of the design for inspection tasks as the pose and field of view of attached sensors will constantly change with the pitch angle. 
Moreover, for pitch angles not equal to zero, the rotor wake directed to the ground leads to high dust generation.

\section{Proposed design: BogieCopter}

\label{section:bogieCopterDesign}

\begin{figure}[tbh]
\centering
\includegraphics[width=0.60\linewidth]{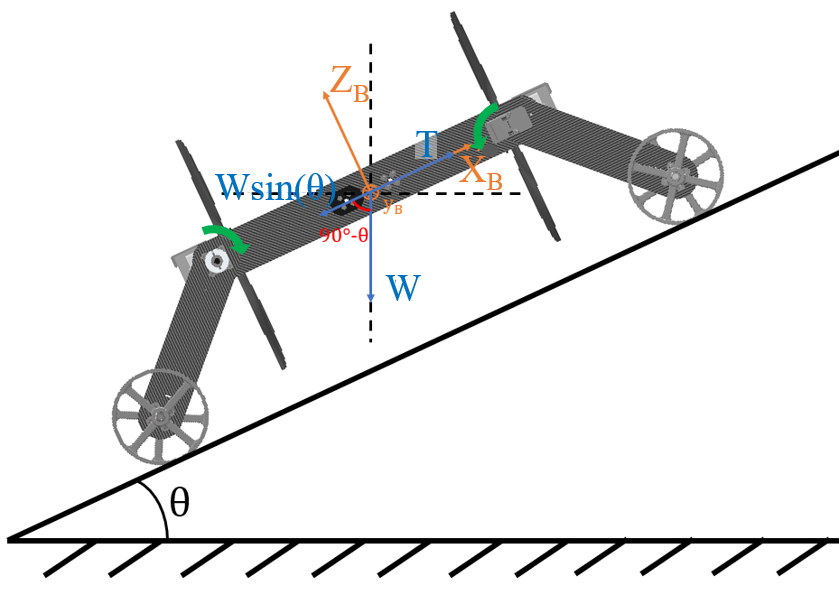}
\vspace{-0.5\baselineskip}
\caption{BogieCopter climbing an inclined surface, showing the rotor position and the forces acting on it (normal force and friction not represented)}
\label{fig:bogieCopterInclined}
\end{figure}

To address the previously mentioned limitations, an innovative passive-actuated ground-aerial vehicle, named BogieCopter, was developed (presented in figure \ref{fig:bogiecopter}). A dual-axle tilt-rotor mechanism was employed to adapt the direction of the rotors based on the mode of locomotion, to have the thrust forces always parallel to the surface while in surface locomotion mode (as shown in  Figure \ref{fig:bogieCopterInclined}), and to point them upwards for aerial locomotion. 
To switch from the flight to ground/surface locomotion mode, the two axles are simply tilted 90$\degree$ inwards, allowing the MAV to quickly accelerate and decelerate longitudinally by controlling the speeds of the rear and front rotors respectively. With rotors in this position, the lateral motion of the vehicle is controlled by the resultant force between the thrust of the left and right rotors. For steep inclinations or wall climbing, both axles are tilted towards the same direction for additional strength.

This passive design is no longer constrained by the assembly of the entire MAV on a central axle, enabling the use of a four-wheeled design which is inherently an stable solution. It allows the decoupling of the ground and the flight controllers and minimizes the dust generation from the propellers when moving on the ground, due to the rotor wake being parallel to the ground. 
By tilting the rotors 90$\degree$, a power efficient surface locomotion is achieved that is independent of the inclination angle. 
Finally, this design allows a stable placement of the payload, such as cameras and other inspection sensors, on a fixed, stable and unobstructed central position. 


\subsection{Design Procedure}


The design of an aerial vehicle is an iterative process due to the interdependency between the many components that compose the vehicle \cite{quadcopterComplete}. BogieCopter's design flow differs from the design of standard multirotors given that additional steps are required for the design and integration of the GAM. 

BogieCopter was designed as a cost-effective platform where all the custom parts can be constructed with a 3D printer and basic workshop tools. It was designed to have a maximum span constrain of 700 mm to be similar to that of commonly available quadrotors used for outdoor industrial applications. 
A T/W ratio of two was considered, based on the rule of thumb used for multirotors \cite{MTMUR, HyTAQ1}, providing a factor of safety for the produced thrust and enabling the platform to be used in windy conditions. The MTOM was estimated totalling 4 kg, accounting 2 kg for the vehicle's mass and 2 kg for the payload (including a SF), the latter one being estimated from the mass of commonly used sensors for inspection applications: RGb, thermal camera and LIDARs.
It was decided that the propulsive system would use single rotors, due to the inefficiency of co-axial rotors \cite{helicopterPrinciples, propellerConfiguration}. For the highest possible efficiency, propellers with the largest possible diameter that respected the vehicle's size constraints were used: APC 10x5E. The propellers analysed for the propulsive system were limited to APC, due to the availability of simulated data for their propellers \cite{apcPerformance}. The motor T-Motor AT2814 was selected based on the simulated characteristics of the propeller and on data from motors' performance charts. Two 4-in-1 ESCs, the Racerstar Air50, are used, given their lower mass and easier assembly when compared with two single ESCs. 

\begin{figure}[h]
\centering
\includegraphics[width=0.875\linewidth]{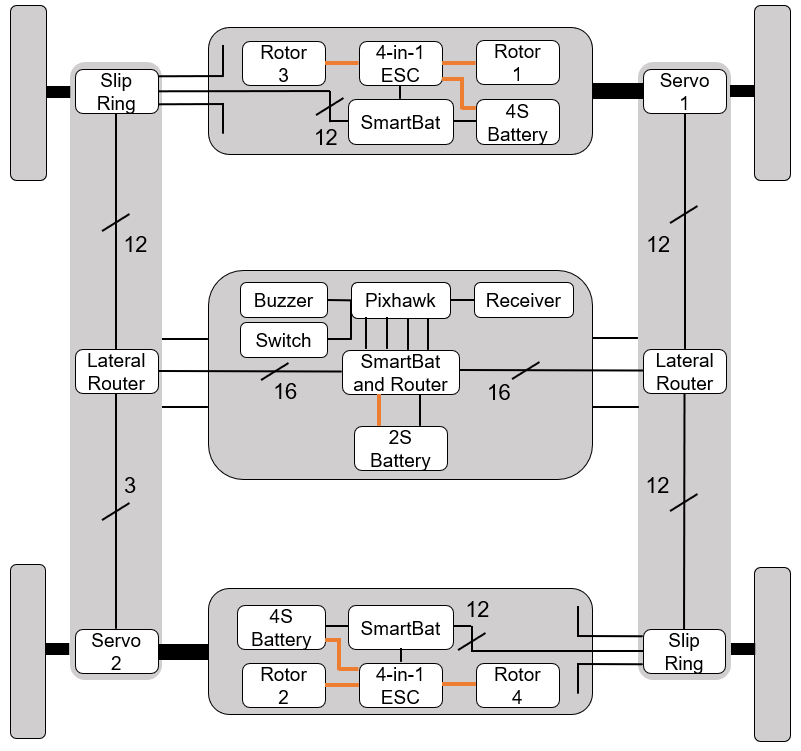}
\vspace{-0.5\baselineskip}
\caption{Hardware and harnessing overview of BogieCopter}
\label{fig:harnessing}
\end{figure}

BogieCopter was designed to achieve a flight time of 10 minutes (based on the estimated power required for the hovering state), with two 4S 4Ah Li-po batteries powering the propulsive system and a small 2s 3.2Ah battery powering the controller and electronics. The use of three independent batteries was necessary due to the modular structure of the design and enable a decoupling between the propulsive's and electronic's power supplies. 
Feetech RC STS3215 servos were selected to control the tilting rotors' arms, providing a torque higher than the gyroscopic moment of the rotating propellers (including a SF) \cite{MTMUR} and full 360º resolution. A Pixhawk 1 autopilot was used as the control unit. 
To accommodate the use of three batteries with the autopilot, two custom made PCBs were designed (the SmartBat and the SmartBat
and Router) which communicate the instantaneous current and individual cell voltages of the batteries through I$^2$C/TWI, and monitor the state of charge (SoC) of the cells, protecting them from over-discharge. A simple harnessing router PCB was also designed, named Lateral Router. An overview of the electronic's design and the harnessing is presented in figure \ref{fig:harnessing}

The mechanical design of BogieCopter was also conceptualised through an iterative process, in which the parts underwent topology optimization and were analysed through static stress analysis. To keep BogieCopter as lightweight as possible, without sacrificing strength, most parts were made out of carbon fibre reinforced plastic (CFRP) tubes and plates. The wheels and non-critical parts were manufactured by using a 3D printer and printed out of ABS. Few critical parts were manufactured from Aluminium 6082. The design also took into consideration the protection of the propellers, where its structure acts as a cage around them. The wheels were designed to have a 50 mm ground clearance, suitable for locomotion on a rocky soil ground. However, other wheel dimensions and designs could be used for different environments and applications, for example, to move on water similar to \cite{MTMUR, ninjaUAV}.
Table \ref{tab:summaryCharacteristicsBogie} summarizes the characteristics of BogieCopter and the components used.

\begin{table}[h]
\centering
\caption{BogieCopter's characteristics and components overview}
\label{tab:summaryCharacteristicsBogie}
\vspace{-0.5\baselineskip}
\begin{threeparttable}
\renewcommand{\arraystretch}{1.1}
\begin{tabular}{cc}
 \hline
\textbf{Body size LxWxH}            & 695x693.5x302 mm\tnote{1}      \\
\hdashline
\textbf{MTOM}                       & 4 kg                   \\
\hdashline
\textbf{Propeller}                  & APC 10x5E              \\
\hdashline
\textbf{Motor}                      & T-Motor AT2814 1050 KV \\
\hdashline
\textbf{ESC}                        & Racerstar Air50        \\
\hdashline
\textbf{Propulsive System Battery}  & 2 x Li-po 4S 5 Ah    \\
\hdashline
\textbf{Electronic's Battery}       & Li-ion 2S 3.2 Ah       \\
\hdashline
\textbf{Servo}                      & Feetech RC STS3215     \\
\hdashline
\textbf{Autopilot}                  & Pixhawk 1              \\
\hdashline
\textbf{Ground Actuation Mechanism} & 4x Custom Made Wheels \\
\bottomrule
\end{tabular}
\begin{tablenotes}
\item [1] Maximum height given with propellers perpendicular to the ground
\end{tablenotes}
\end{threeparttable}
\end{table}

\section{Experiments and Results}
\label{section:results}

\begin{figure}[h]
\centering
\includegraphics[width=1\linewidth]{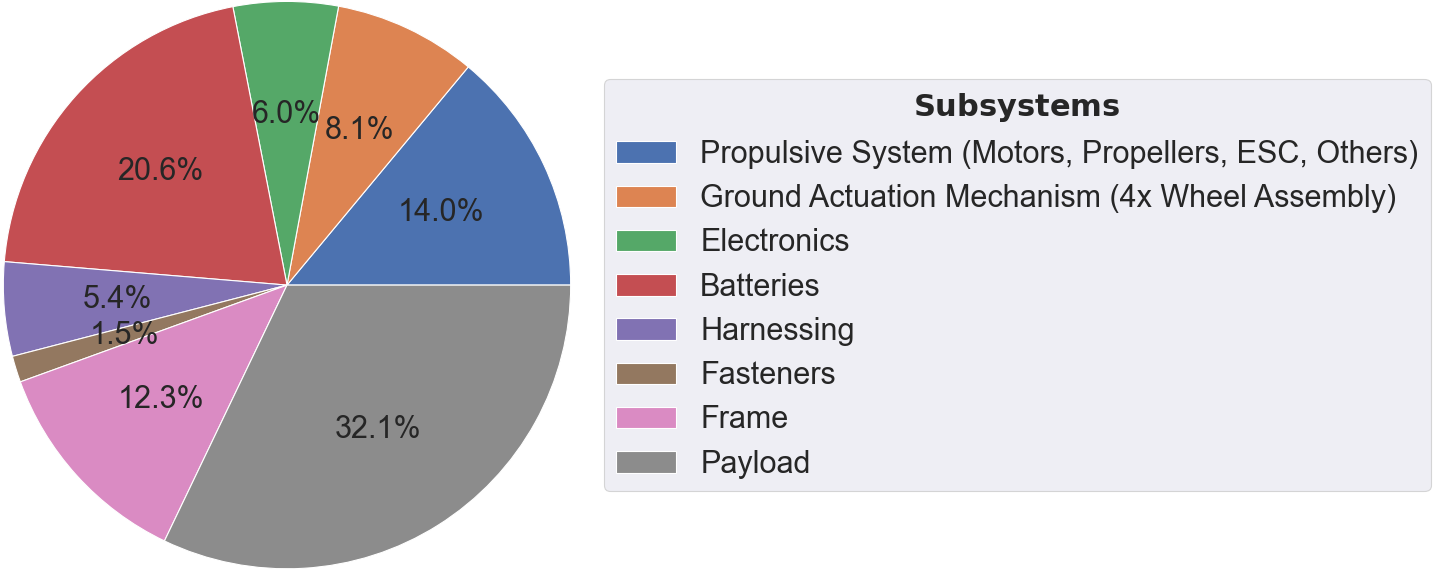}
\caption{Overview of BogieCopter's mass distribution for its main subsystems}
\label{fig:massDistro}
\vspace{-0.75\baselineskip}
\end{figure}

The mass of BogieCopter without payload was determined to be 2.7 kg $\pm$ 0.1 kg, as presented in figure \ref{fig:massDistro}. This value was higher than the initial design target, caused by the over-dimensioning of some of the manufactured CFRP components. 
Still considering a MTOM of 4 kg, this implies that BogieCopter can carry a payload of 1.3 kg $\pm$ 0.1 kg. The GAM has a total mass of 328.8 g $\pm$ 0.4 g, corresponding to 8.2\% of the MTOM. To the best of our knowledge, this is the lowest GAM's mass percentage for a multi-modal MAV that has the ability to remain stationary without actuation. 

\begin{figure}[h]
\centering
\includegraphics[width=1\linewidth]{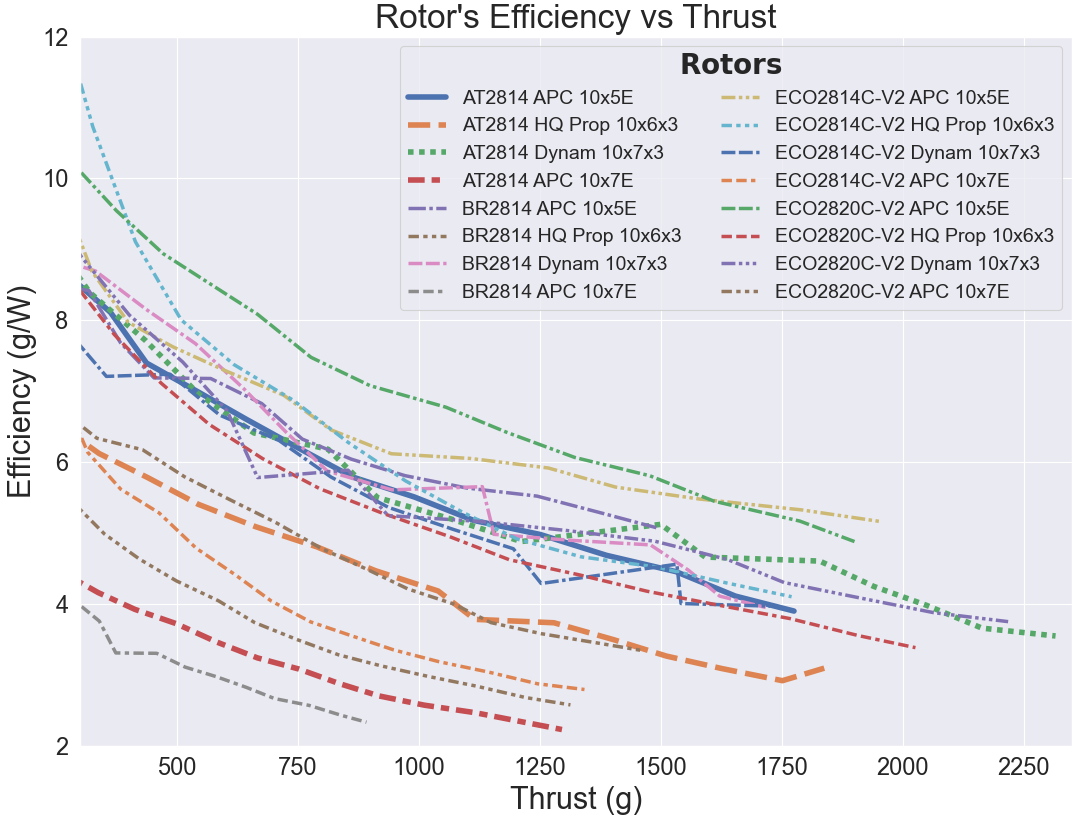}
\vspace{-1.75\baselineskip}
\caption{Comparison between the efficiency vs thrust of different rotors, obtained from the static thrust tests realised in a thrust stand}
\label{fig:efficiencyThrust}
\end{figure}

Static thrust tests were initially conducted in a thrust stand, in order to validate the selected rotor. These tests demonstrated that the rotor was able to produce a maximum thrust of 1.843 kg, 
resulting in a MAV's T/W ratio (for the MTOM) of 1.843. 
To the best of our knowledge this is the best T/W ratio among other designs that have a payload capacity.
The selected rotor was also compared against fifteen rotors (created from four different propellers and four different motors). 
Figure \ref{fig:efficiencyThrust} presents the efficiency in relation to thrust for all the tested rotors. It can be seen that, by replacing the selected motor to the Dualsky ECO2814C-V2, the operating time of the vehicle can be increased further due to the higher efficiency for the same propeller.

\begin{figure}[h!]
\centering

\hspace*{\fill}
\subfigure{\label{fig:flight1}
\xincludegraphics[width=0.275\linewidth,label=\subref{fig:flight1},labelbox]{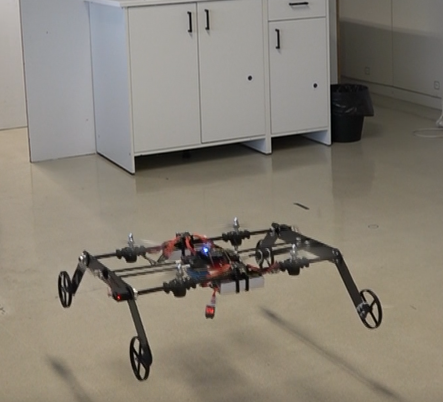}}
\hfill
\subfigure{\label{fig:flight2}
\xincludegraphics[width=0.275\linewidth,label=\subref{fig:flight2},labelbox]{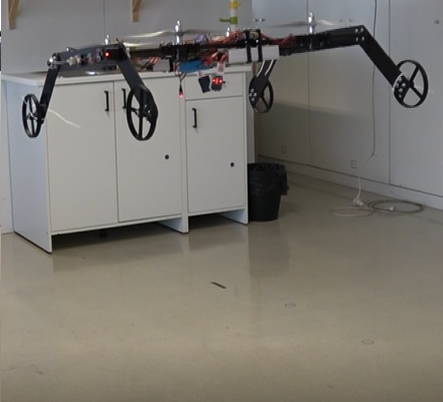}}
\hfill
\subfigure{\label{fig:flight3}
\xincludegraphics[width=0.275\linewidth,label=\subref{fig:flight3},labelbox]{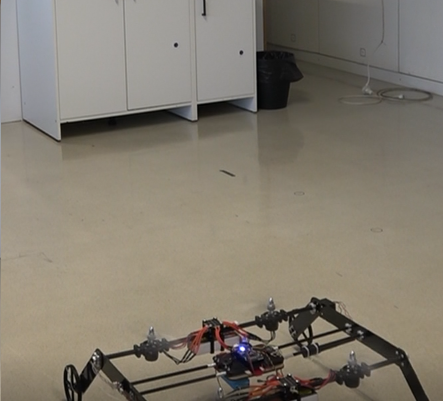}}
\hspace*{\fill}

\vspace{-0.5\baselineskip}
\caption{BogieCopter flight test:
\subref{fig:flight1} Climbing
\subref{fig:flight2} Hovering
\subref{fig:flight3} Landed}
\label{fig:bogieFlying}
\end{figure}

A set of real-world experiments\footnote{Video highlights available at \url{https://youtu.be/ZeZAYKi3F2s}} were executed to test and validate BogieCopter for all modes of locomotion: 

\textbf{Aerial Locomotion:} The first experiments performed were flight tests, using the vastly tested ArduCopter firmware, creating a performance baseline for the vehicle and enabling the assessment of the vehicle's construction. Since the initial tests, BogieCopter demonstrated a very stable and normal flight behaviour. Figure \ref{fig:bogieFlying} shows instances from a flight test.

\begin{figure}[h!]
\centering
\begin{minipage}[h]{1.0\linewidth}
\centering
\hspace*{\fill}
\subfigure{\label{fig:confined1}
\xincludegraphics[width=0.3\linewidth,label=\subref{fig:confined1},labelbox]{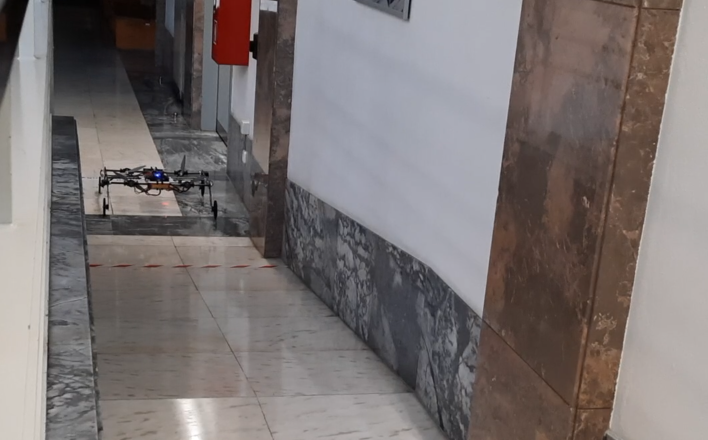}}
\hfill
\subfigure{\label{fig:confined2}
\xincludegraphics[width=0.3\linewidth,label=\subref{fig:confined2},labelbox]{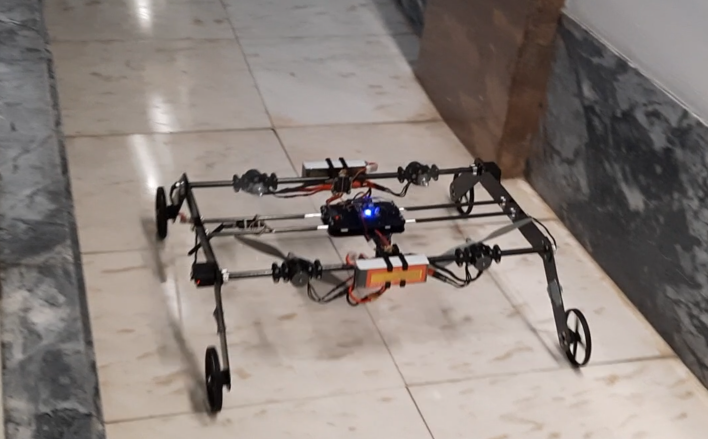}}
\hfill
\subfigure{\label{fig:confined3}
\xincludegraphics[width=0.3\linewidth,label=\subref{fig:confined3},labelbox]{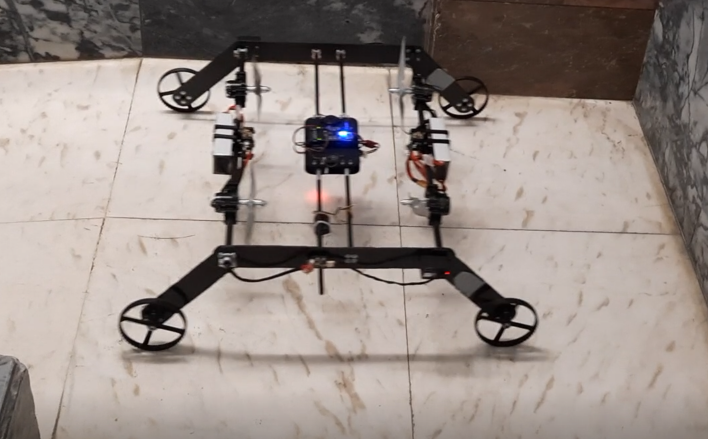}}
\hspace*{\fill}
\end{minipage}

\vspace{-7.5pt}

\begin{minipage}[h]{1.0\linewidth}
\centering
\hspace*{\fill}
\subfigure{\label{fig:confined4}
\xincludegraphics[width=0.3\linewidth,label=\subref{fig:confined4},labelbox]{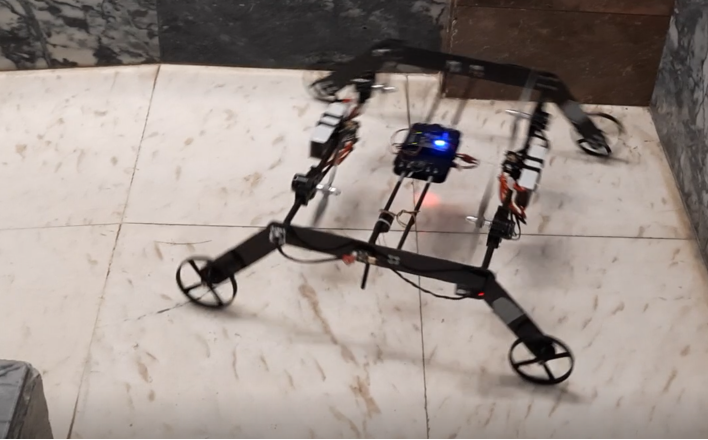}}
\hfill
\subfigure{\label{fig:confined5}
\xincludegraphics[width=0.3\linewidth,label=\subref{fig:confined5},labelbox]{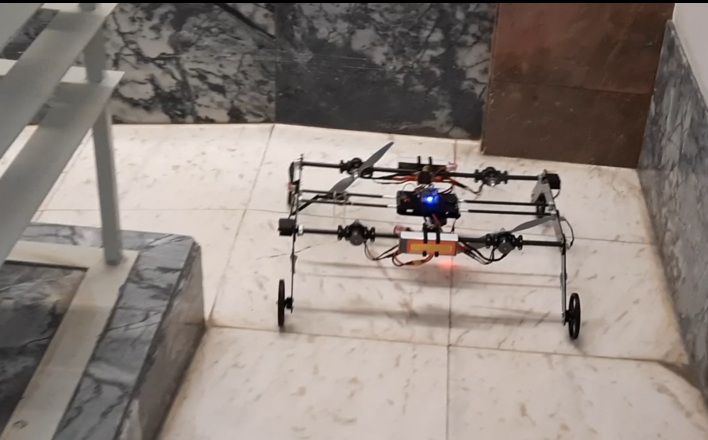}}
\hfill
\subfigure{\label{fig:confined6}
\xincludegraphics[width=0.3\linewidth,label=\subref{fig:confined6},labelbox]{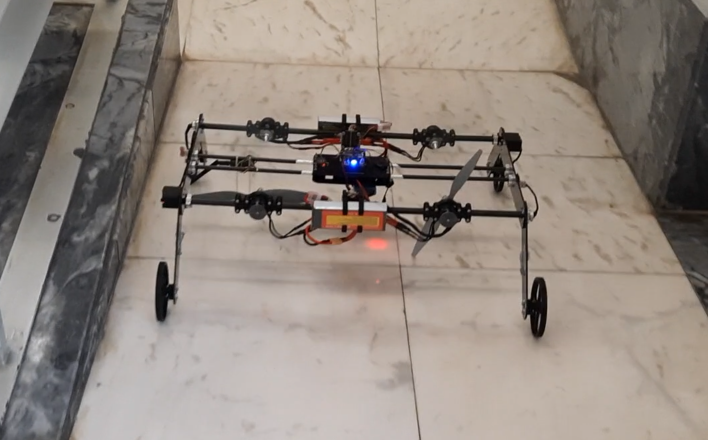}}
\hspace*{\fill}
\end{minipage}
\vspace{-1\baselineskip}
\caption{Test on a long narrow wheel-chair ramp (simulated confined space:
\subref{fig:confined1} Start of the test
\subref{fig:confined2} Moving longitudinally
\subref{fig:confined3} Arriving at a corner
\subref{fig:confined4} - \subref{fig:confined5} Turning
\subref{fig:confined6} Exiting the ramp}
\label{fig:confinedSpace}
\end{figure}

\begin{figure}[h!]
\centering

\hspace*{\fill}
\subfigure{\label{fig:rocky1}
\xincludegraphics[width=0.3\linewidth,label=\subref{fig:rocky1},labelbox]{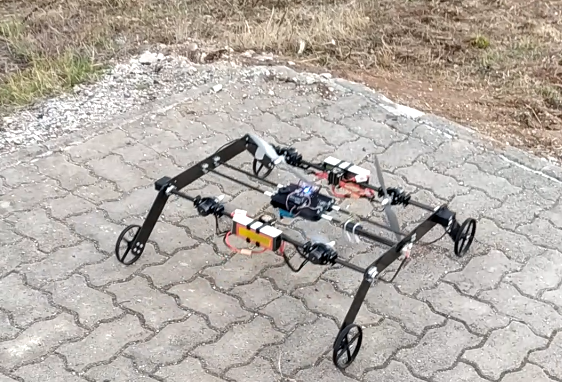}}
\hfill
\subfigure{\label{fig:rocky2}
\xincludegraphics[width=0.3\linewidth,label=\subref{fig:rocky2},labelbox]{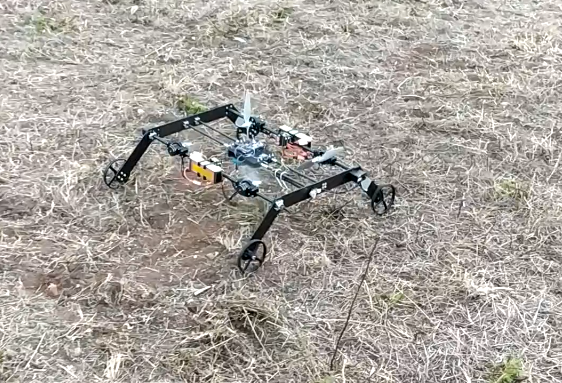}}
\hfill
\subfigure{\label{fig:rocky3}
\xincludegraphics[width=0.3\linewidth,label=\subref{fig:rocky3},labelbox]{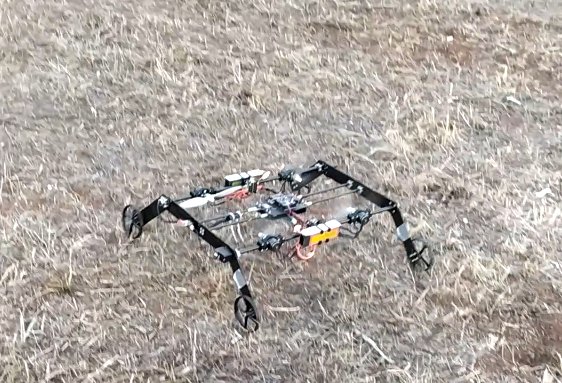}}
\hspace*{\fill}

\vspace{-0.5\baselineskip}
\caption{Test on a rocky uneven soil ground:
\subref{fig:rocky1} Start of the test in a sidewalk
\subref{fig:rocky2} Moving on the rocky soil ground, moving over grass
\subref{fig:rocky3} Turning on the rocky soil ground}
\label{fig:rockySoil}
\end{figure}

\textbf{Flat Surface Locomotion:} Having concluded successfully the flight tests, the performance of BogieCopter to move on flat surfaces was extensively evaluated. These tests included simple longitudinal and lateral motion tests up to more complex and challenging routes. Figure \ref{fig:confinedSpace} illustrates BogieCopter moving in a narrow space, with the vehicle having a clearance of only 0.15 m to the walls on either side of it. While it would be almost impossible to fly a MAV in this narrow space, without the possibility of collision with the walls, the ground mobility capability allowed it to successfully perform the task. Other tests were conducted that also demonstrated the ground locomotion capability in rocky and uneven ground, as presented in figure \ref{fig:rockySoil}.

\begin{figure}[h!]
\centering

\begin{minipage}[h]{1.0\linewidth}
\centering
\hspace*{\fill}
\subfigure{\label{fig:modal1}
\xincludegraphics[width=0.425\linewidth,label=\subref{fig:modal1},labelbox]{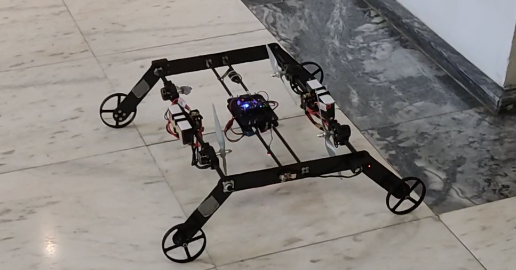}}
\hfill
\subfigure{\label{fig:modal2}
\xincludegraphics[width=0.425\linewidth,label=\subref{fig:modal2},labelbox]{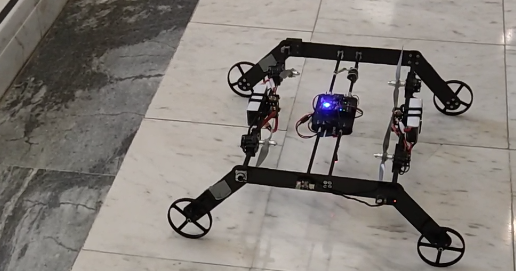}}
\hspace*{\fill}
\end{minipage}

\vspace{-7.5pt}

\begin{minipage}[h]{1.0\linewidth}
\centering
\hspace*{\fill}
\subfigure{\label{fig:modal3}
\xincludegraphics[width=0.425\linewidth,label=\subref{fig:modal3},labelbox]{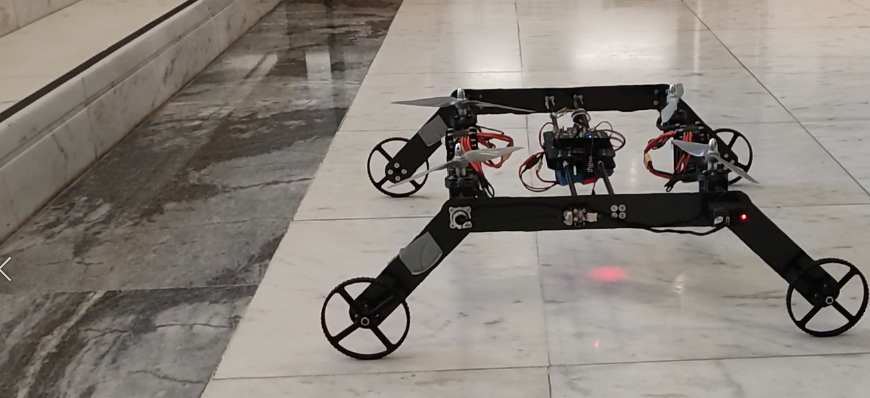}}
\hfill
\subfigure{\label{fig:modal4}
\xincludegraphics[width=0.425\linewidth,label=\subref{fig:modal4},labelbox]{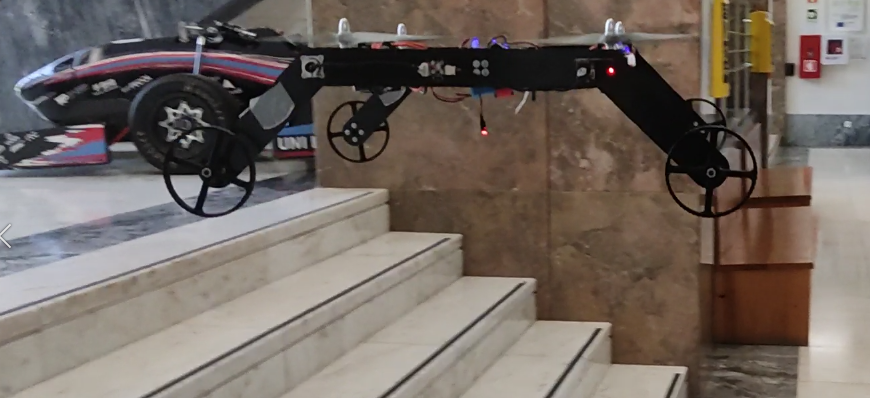}}
\hspace*{\fill}
\end{minipage}

\vspace{-7.5pt}

\begin{minipage}[h]{1.0\linewidth}
\centering
\hspace*{\fill}
\subfigure{\label{fig:modal5}
\xincludegraphics[width=0.425\linewidth,label=\subref{fig:modal5},labelbox]{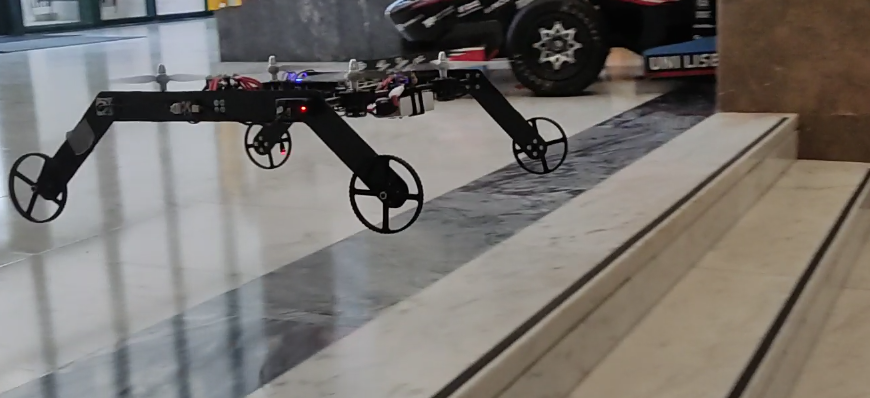}}
\hfill
\subfigure{\label{fig:modal6}
\xincludegraphics[width=0.425\linewidth,label=\subref{fig:modal6},labelbox]{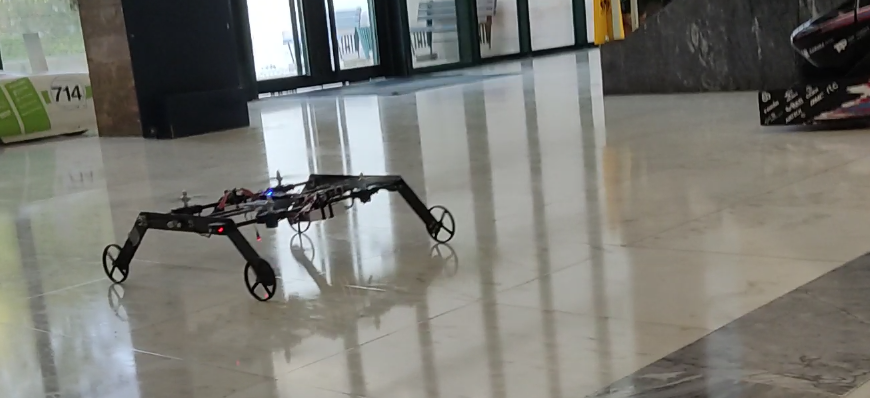}}
\hspace*{\fill}
\end{minipage}

\vspace{-0.5\baselineskip}
\caption{Demonstration of the multi-modal capabilities of BogieCopter:
\subref{fig:modal1} Start of the ground locomotion
\subref{fig:modal2} MAV facing an obstacle
\subref{fig:modal3} Change to flight mode
\subref{fig:modal4} - \subref{fig:modal5} Flying to overcome obstacle
\subref{fig:modal6} Once obstacle is surpassed, change back to more efficient ground mode}
\label{fig:multiModal}
\end{figure}

\textbf{Multi-modal Capability:} Having established the flight and ground locomotion performance of BogieCopter, tests exploring the multi-modality of the vehicle were performed. In these, the vehicle, when moving on the ground, has to overcome an obstacle (fly over it), to proceed with its path. A set of images representing one of these tests are depicted in figure \ref{fig:multiModal}.

\begin{figure}[h!]
\centering

\hspace*{\fill}
\subfigure{\label{fig:inclined1}
\xincludegraphics[width=0.4\linewidth,label=\subref{fig:inclined1},labelbox]{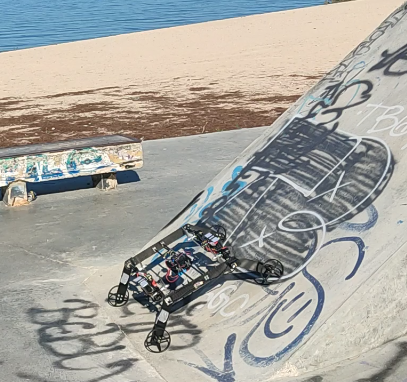}}
\hfill
\subfigure{\label{fig:inclined2}
\xincludegraphics[width=0.4\linewidth,label=\subref{fig:inclined2},labelbox]{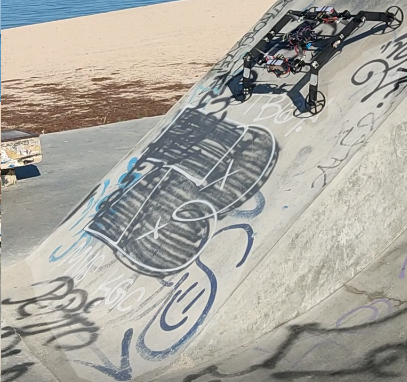}}
\hspace*{\fill}

\vspace{-0.5\baselineskip}
\caption{Test on 33$\degree$ inclined surface:
\subref{fig:inclined1} Start of the climb
\subref{fig:inclined2} Reaching the top of the inclined surface}
\label{fig:inclinedSurface}
\end{figure}

\textbf{Inclined Surface Locomotion:} One of the limitations in the previous multi-modal MAVs works was the assessment of the vehile's ability to move on inclined surfaces, having been only demonstrated in \cite{ninjaUAV}, through a stars' climb test. Due to BogieCopter's design, the behaviour of the MAV on inclined surfaces (up to 60.93$\degree$ slopes, from which the vehicle can start to tip) is similar to the one that the vehicle exhibits when moving on flat surfaces. Furthermore, \textit{a priori} knowledge of the slope angle isn't required for having the efficient locomotion. BogieCopter's ability to move on inclined surfaces was validated through multiple tests, from low slope angles of 2$\degree$ up to medium slope angles of 33$\degree$. Figure \ref{fig:inclinedSurface} shows BogieCopter climbing a 33$\degree$ inclined ramp in a skate park.


\begin{figure}[h!]
\centering

\hspace*{\fill}
\subfigure{\label{fig:wallClimb1}
\xincludegraphics[width=0.20\linewidth,label=\subref{fig:wallClimb1},labelbox]{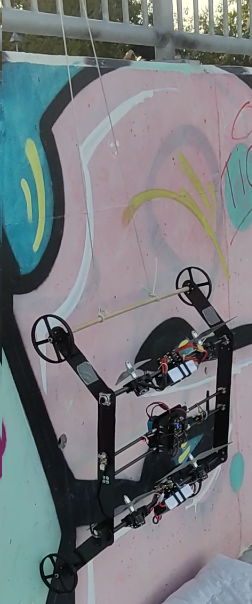}}
\hfill
\subfigure{\label{fig:wallClimb2}
\xincludegraphics[width=0.20\linewidth,label=\subref{fig:wallClimb2},labelbox]{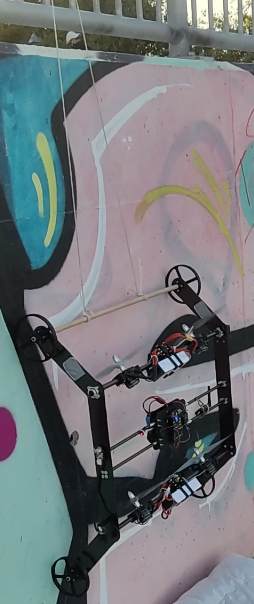}}
\hfill
\subfigure{\label{fig:wallClimb3}
\xincludegraphics[width=0.20\linewidth,label=\subref{fig:wallClimb3},labelbox]{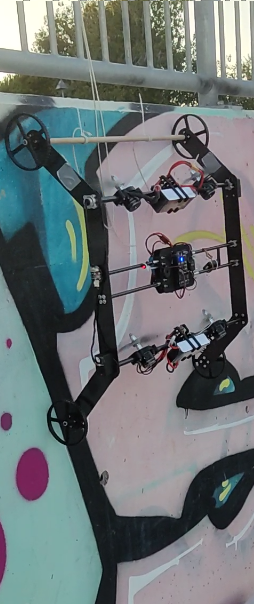}}
\hspace*{\fill}

\vspace{-0.5\baselineskip}
\caption{Test the capabilities of BogieCopter to wall-climb:
\subref{fig:wallClimb1} Developed MAV being suspended close to a wall by a rope (rope in tension)
\subref{fig:wallClimb2} MAV sticking to the wall, due to the downforce created by the rotors
\subref{fig:wallClimb3} MAV climbing the wall (notice that the rope is no longer in tension}
\label{fig:wallClimb}
\end{figure}

\textbf{Wall-Climbing:} BogieCopter's design should enable it to wall-climb, similarly to \cite{wallClimber}. This ability can have benefits for several applications \cite{wallClimber}, such us for contact-based inspection, opening up a new range of possibilities for the developed MAV. A test was conducted to validate this assumption, as depicted in figure \ref{fig:wallClimb}. This test was conducted with the rotors tilted at an angle of 135$\degree$, to remain attached and climb the wall.

\begin{figure}[tbh]
\centering
\includegraphics[width=1\linewidth]{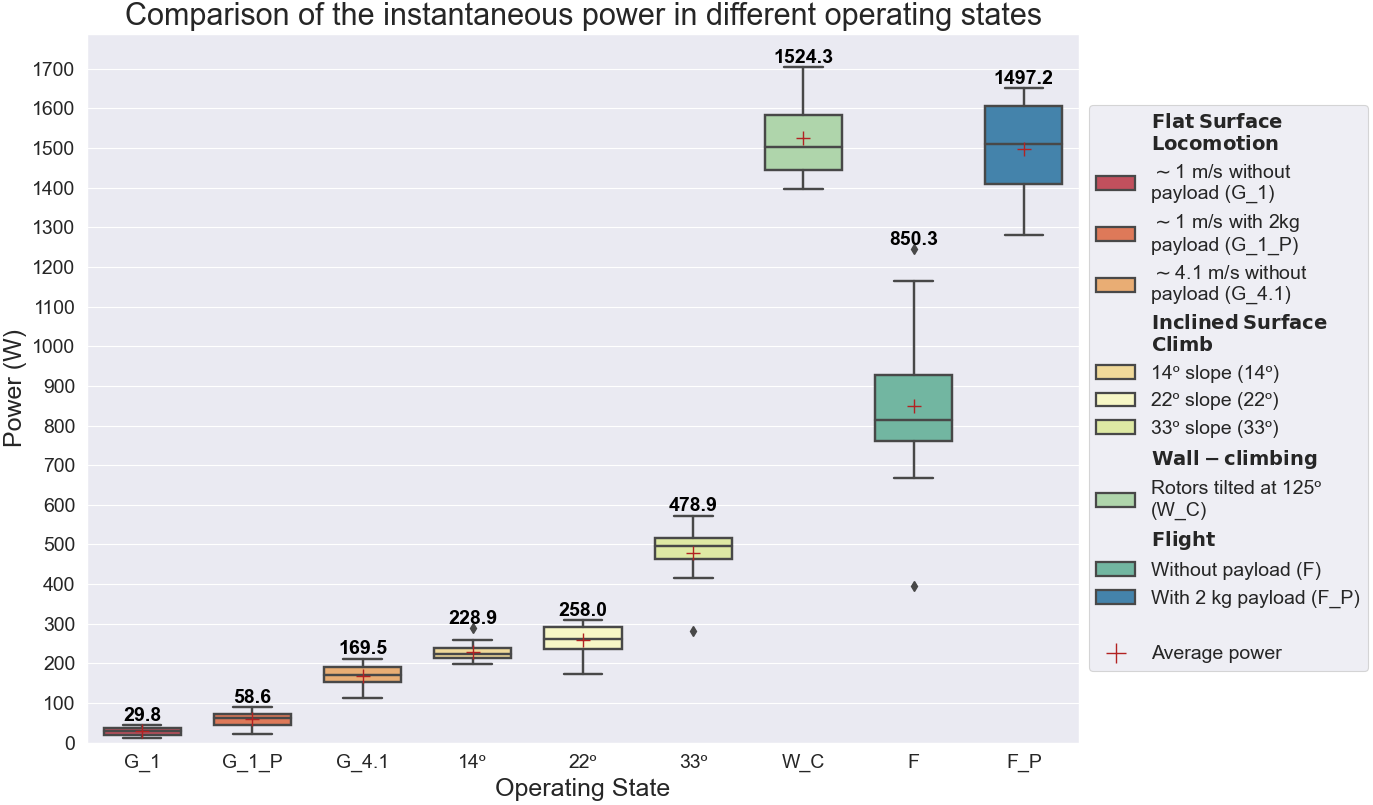}
\vspace{-1.75\baselineskip}
\caption{Comparison between the power required to maintain each operating state. The value above the box plots corresponds to the average power consumption}
\label{fig:summaryPlot}
\end{figure}

During all tests, the instantaneous current and battery voltage were logged, enabling comprehensive data for analyzing the power demand of each operating state. Figure \ref{fig:summaryPlot} presents the power consumption across all modes of operation.
As expected, the ground locomotion on flat surfaces is the state consuming the least energy. In fact, comparing the flight without payload to the ground locomotion without payload, with the same speed of 1 m/s, the ground locomotion requires 28.8x less energy than the flight, only consuming 29.8 W. This is the lowest value among the works that shared their power's details, being even better than the power consumption of the active actuated designs. The locomotion modes with a 2 kg payload require, approximately, 2 times more energy than the correspondent modes without payload. Moving on the ground with a 2 kg payload at 1 m/s, only requires 58.6 W, 25.5x less than flying with the payload. Considering the batteries currently being used in BogieCopter, the vehicle is able to achieve an operating distance of $\sim$ 11.5 km while moving on flat surfaces at 1 m/s and $\sim$ 8.2 km while moving at 4.1 m/s. 

As expected, the power required to remain on an inclined surface increases with the inclination angle, being more efficient than a hovering state. 
An unforeseen power requirement arises during wall-climbing, which is significantly higher than flying. This can be attributed to the rotor tilt angle, which is larger than the optimal angle required to keep the vehicle attached to the wall without tipping over. Additionally, the interference of the wake from the front rotors on the rear rotors, owing to the high rotor speeds required for climbing, could be another contributing factor.




\section{Conclusion and Future Work}
\label{section:conclusion}

This paper presents the design, development and testing of a novel passive actuated multi-modal MAV. 
The design allows a generous payload capacity, suitable for many industrial applications, while offering extended operating times due to its multi-modality and efficient passive ground actuation mechanism (GAM). The design was validated through a set of real-world experiments, and for different environments, demonstrating flying, ground and inclined surface locomotion, and wall-climbing. To the best of our knowledge, BogieCopter has the lowest percentage of mass for the GAM, relative to MTOM, considering all designs that don't require actuation to be stationary. It has the highest payload capacity, considering the same T/W ratio for all the prior designs, and it achieves the lowest power consumption for ground locomotion among known works. 
Improving the operating time through optimization of the CFRP components, identifying the optimal material and size for the wheels, developing position controllers for the wall climbing and surface locomotion modes, and improvement of the flight mode through exploiting the tilting rotor mechanism are some of the work we are currently pursuing.  






\bibliographystyle{IEEEtran}
\bibliography{IEEEabrv,biblio}


\end{document}